\documentclass[psamsfonts,oneside]{amsart}
\usepackage[foot]{amsaddr}
\usepackage{cmap}
\usepackage[utf8x]{inputenc}
\usepackage[T1]{fontenc}
\usepackage[english]{babel}
\usepackage{graphicx}
\usepackage{latexsym}
\usepackage{amsmath}
\usepackage{amssymb}
\usepackage{amsthm}
\newtheorem{thm}{Theorem}[section]

\theoremstyle{definition}

\usepackage{cite}
\usepackage[colorlinks=true,linkcolor=blue]{hyperref}

\providecommand{\norm}[1]{\lVert#1\rVert}
\begin{document}

\title[Akimoto et al.: In Algorithmica, Online First (2011)]{Theoretical Foundation for CMA-ES from Information Geometry Perspective}

\author{Youhei Akimoto}
\address[Y.~Akimoto]{TAO Team - INRIA Saclay, LRI - Paris-Sud University 91405 Orsay, France}
\email{Youhei.Akimoto@lri.fr}

\author{Yuichi Nagata}
\author{Isao Ono}
\author{Shigenobu Kobayashi}
\address[Y.~Nagata, I.~Ono, S.~Kobayashi]{Interdisciplinary Graduate School of Science and Engineering, Tokyo Institute of Technology 226-8502 Kanagawa, Japan}
\date{}
\maketitle

\begin{abstract}
%This paper explores the theoretical basis of the covariance matrix adaptation evolution strategy (CMA-ES) from the information geometry viewpoint. To establish a theoretical foundation, we consider the objective of the distribution parameter update to be maximization of the expected fitness under the sampling distribution. We investigate the steepest ascent learning for the expected fitness maximization, where the steepest ascent direction is given by the natural gradient, which is the product of the inverse of the Fisher information matrix and the conventional gradient of the function. Our first result is that we can obtain the natural gradient of the expected fitness without the need for inversion of the Fisher information matrix under some parameterization. We find that the update of the distribution parameters in the CMA-ES is the same as natural gradient learning for expected fitness maximization. Second, we derive the range of learning rates such that a step in the direction of the exact natural gradient improves the parameters in the expected fitness. We see from the close relation between the CMA-ES and natural gradient learning that the default setting of learning rates in the CMA-ES seems suitable in terms of monotone improvement in expected fitness. Finally, we discuss the relation to the expectation-maximization framework and provide an information geometric interpretation of the CMA-ES.
This paper explores the theoretical basis of the covariance matrix adaptation evolution strategy (CMA-ES) from the information geometry viewpoint.

 To establish a theoretical foundation for the CMA-ES, we focus on a geometric structure of a Riemannian manifold of probability distributions equipped with the Fisher metric. We define a function on the manifold which is the expectation of fitness over the sampling distribution, and regard the goal of update of the parameters of sampling distribution in the CMA-ES as maximization of the expected fitness. We investigate the steepest ascent learning for the expected fitness maximization, where the steepest ascent direction is given by the natural gradient, which is the product of the inverse of the Fisher information matrix and the conventional gradient of the function.

 Our first result is that we can obtain under some types of parameterization of multivariate normal distribution the natural gradient of the expected fitness without the need for inversion of the Fisher information matrix. We find that the update of the distribution parameters in the CMA-ES is the same as natural gradient learning for expected fitness maximization. Our second result is that we derive the range of learning rates such that a step in the direction of the exact natural gradient improves the parameters in the expected fitness. We see from the close relation between the CMA-ES and natural gradient learning that the default setting of learning rates in the CMA-ES seems suitable in terms of monotone improvement in expected fitness. Then, we discuss the relation to the expectation-maximization framework and provide an information geometric interpretation of the CMA-ES.
\end{abstract}
\newpage
\tableofcontents
{\small 
This article appears in \textit{Algorithmica Journal}, DOI: 10.1007/s00453-011-9564-8.\\

\paragraph*{An erratum} 
In the definition of $Q(\theta, \theta')$ (that is above \eqref{eq:lower-bound} in Section~\ref{sec:improvement}) there was a ``$-$'' in front of the integral sign on the right-most side which should not be there. This is corrected in this version.
}

\newpage
% Section 1 %%%%%%%%%%%%%%%%%%%%%%%%%%%%%%%%%%%%%%%%%%%%%%%%%%%%%%%%%%%%%%%%%%%
\section{Introduction}
\label{sec:intro}

The covariance matrix adaptation evolution strategy (CMA-ES; e.g., \cite{HansenEC2001,HansenEC2003}) is the leading stochastic and derivative-free algorithm for solving continuous optimization problems, i.e., for finding the optimizer $\mathbf{x}^*$ of a real-valued objective function $f$, aka fitness, defined on (a subset of) $\mathbb{R}^d$, which we assume to be maximized without loss of generality. The CMA-ES generates candidate points $\{\mathbf{x}_{i}\}$, $i \in \{1, 2, \dots, \lambda\}$, from a multivariate normal distribution and evaluates their fitness values $\{f(\mathbf{x}_{i})\}$. Then, it updates the mean vector and covariance matrix of the multivariate normal distribution by using the information of the sampled points and their fitness values, $\{(\mathbf{x}_{i}, f(\mathbf{x}_{i}))\}$. Repeating the sampling-evaluation-update procedure, the CMA-ES moves the sampling distribution to a promising area over and over, and is expected to find a neighborhood of the optimizer. At least, we do not expect it to converge to a non-stationary point of the objective function \cite{AkimotoGECCO2010}.

The method used to improve the parameters of the sampling distribution strongly determines the behavior and efficiency of the whole algorithm. The CMA-ES updates the parameters so that it encourages to reproduce previously successful search steps. To do so, the CMA-ES, especially the rank-$\mu$ update in the CMA-ES \cite{HansenEC2003} is based on a maximum-likelihood estimation. Hence, the CMA-ES can be considered to be based on a statistical principle.

Recently, Wierstra et al.~\cite{WierstraCEC2008} proposed a novel algorithm named natural evolution strategy (NES), which was subsequently developed further by Sun et al.~\cite{SunGECCO2009,SunICML2009} and Glasmachers et al.~\cite{GlasmachersGECCO2010}. In NESs, the objective of the parameter update is considered to be maximization of the expected fitness $\mathbb{E}[f(\mathbf{x})]$, where the expectation is taken under the current sampling distribution, and a natural gradient \cite{AmariNC1998} based approach is employed. Thus, NESs are considered to be derived from a principle of information geometry and, from their nature, constitute a more principled approach than the CMA-ES.

This paper addresses the theoretical justification for the CMA-ES from the information geometry viewpoint and gives a mathematical interpretation of the CMA-ES. For this purpose, we consider a geometric structure of a Riemannian manifold of probability distributions equipped with the Fisher metric, and define an alternate maximization problem on the manifold: the objective function is the expectation $\mathbb{E}[f(\mathbf{x})\mid \theta]$ of the fitness function, where the expectation is taken under the normal distribution parameterized by $\theta$, and the arguments are the parameters $\theta$ of the normal distribution. Then, we investigate natural gradient learning, i.e.\ steepest ascent learning on the manifold, for the expected fitness maximization. This idea is thoroughly inspired by the formulation of NESs.

The first result of this paper is an analogy between the CMA-ES and natural gradient learning for expected fitness maximization. We show that the natural gradient, which is given by the product of the inverse of the Fisher information matrix of the normal distribution and the conventional gradient, can be directly estimated without calculation of the Fisher information matrix and its inverse under some particular parameterization of the normal distribution. Then, we see that the natural gradient learning for maximizing the expected fitness where the natural gradient is estimated from the samples in a particular parameterization, has the same form of parameter update as the CMA-ES. This part of the paper is the extension of our previous study \cite{AkimotoPPSN2010}.

The second part of this article deals with the learning rate parameter. The natural gradient view of the CMA-ES gives us an insight into the learning rate: the learning rate does not only possess an effect of reducing fluctuation of the parameters due to the variance of the natural gradient estimate, but also takes control of the step-size along with the natural gradient. In a general scheme of gradient-based learning, scheduling of the learning rate is an important factor in determining the speed and accuracy of convergence and the optimal learning rate varies with the function and the position of the parameter \cite{AmariNC1998}. However, the learning rates in the CMA-ES are usually fixed during learning and they are different for the mean vector and for the covariance matrix. Here, an interesting question arises as to why the CMA-ES performs well with constant learning rates within $(0, 1]$ that are different for each parameter. To confirm the validity of this setting, we derive the range of learning rates which guarantee that a step along the exact natural gradient improves the expected fitness value. Then, we discuss the similarity to the fitness expectation-maximization algorithm \cite{WierstraPPSN2008} which is based on expectation-maximization (EM; \cite{DempsterJRSS1977}) framework, and provide an information geometric interpretation of the CMA-ES as natural gradient learning for expected fitness maximization.

The rest of this paper is organized as follows: Section~\ref{sec:cma} introduces the CMA-ES. Section~\ref{sec:ngl} introduces the framework of natural gradient learning for expected fitness maximization. Section~\ref{sec:natural} derives the form of the natural gradient estimate and shows that the CMA-ES and natural gradient learning for expected fitness maximization have the same form of parameter update and that we can describe the CMA-ES and NESs using the same framework. Section~\ref{sec:improvement} provides the range of learning rates so that exact natural gradient learning leads to monotone improvement in the expected fitness, followed by a discussion about the learning rates in the CMA-ES. We discuss the relation to the EM-inspired algorithm \cite{WierstraPPSN2008} and the correspondence to the framework of generalized EM (GEM) algorithms \cite{DempsterJRSS1977}. We conclude with a summary in Section~\ref{sec:summary}. 

% Section 2 %%%%%%%%%%%%%%%%%%%%%%%%%%%%%%%%%%%%%%%%%%%%%%%%%%%%%%%%%%%%%%%%%%%
\section{Covariance Matrix Adaptation Evolution Strategy}
\label{sec:cma}

Let $\pi(\mathbf{x}; \mathbf{m}, \sigma^2\mathbf{C})$ represent the probability density function of the multivariate normal distribution with mean vector $\mathbf{m}$ and covariance matrix $\sigma^2\mathbf{C}$. Here $\sigma$ is a scalar and we call $\sigma$ a global step-size in the context of CMA-ES. The CMA-ES \cite{Hansen2006inbook} repeats the following steps after choosing the initial parameters $\mathbf{m}^0$, $\sigma^0$ and $\mathbf{C}^0$ and setting $\mathbf{p}_\sigma^0 = \mathbf{0}$ and $\mathbf{p}_C^0 = \mathbf{0}$.
\begin{enumerate}
 \item Sample $\lambda$ independent points $\mathbf{x}_1, \dots, \mathbf{x}_\lambda$ from $\pi\left(\mathbf{x}; \mathbf{m}^t, (\sigma^t)^2 \mathbf{C}^t\right)$.
 \item Evaluate the fitness values $f(\mathbf{x}_1), \dots, f(\mathbf{x}_\lambda)$.
 \item Update the parameters as follows.
\begin{description}
 \item[\textit{Mean vector}:] 
\begin{equation*}
 \mathbf{m}^{t+1} = \sum_{i=1}^\lambda \mathrm{w}_{R_i} \mathbf{x}_{i},
\end{equation*}
where $R_i$ represents the ranking of $f(\mathbf{x}_i)$, i.e., $\mathbf{x}_i$ has the $R_i^\text{th}$ highest fitness value among $f(\mathbf{x}_1), \dots, f(\mathbf{x}_\lambda)$; and $\mathrm{w}_{R_i}$ represents the weight for the $R_i^\text{th}$ highest point and has the following properties: $0 \leq \mathrm{w}_{i} \leq \mathrm{w}_j \leq 1$ for any $i > j$ and $\sum_{i=1}^\lambda \mathrm{w}_{i} = 1$. 
 \item[\textit{Global step-size}:] 
\begin{equation*}
 \sigma^{t+1} = \sigma^t \exp\left(\frac{c_\sigma}{d_\sigma} \frac{\norm{\mathbf{p}_{\sigma}^{t+1}} - \chi_d}{\chi_d}\right),
\end{equation*}
where $c_\sigma$ and $d_\sigma$ are the learning rate and the damping parameter, respectively; $\chi_d$ denotes the expectation of the chi distribution with $d$ degrees of freedom; $\mathbf{p}_\sigma$ is an evolution path that is updated as\begin{equation*}
 \mathbf{p}_\sigma^{t+1} = (1-c_\sigma) \mathbf{p}_\sigma^{t} + \sqrt{\frac{c_\sigma(2-c_\sigma)}{\sum_{i=1}^\lambda \mathrm{w}_{i}^{2}}} \frac{(\mathbf{C}^{t})^{-1/2}(\mathbf{m}^{t+1}-\mathbf{m}^t)}{\sigma^{t}}.
\end{equation*}
\item[\textit{Covariance matrix}:] 
\begin{equation*}
\mathbf{C}^{t+1} = (1-c_1-c_\mu) \mathbf{C}^t + c_1 \mathbf{p}_C^{t+1}(\mathbf{p}_C^{t+1})^\mathrm{T} + c_\mu \sum_{i=1}^\lambda \mathrm{w}_{R_i} \frac{\mathbf{x}_{i} - \mathbf{m}^{t}}{\sigma^t} \left(\frac{\mathbf{x}_{i} - \mathbf{m}^{t}}{\sigma^t}\right)^\mathrm{T},
\end{equation*}
where $c_1$ and $c_\mu$ are learning rate parameters and $\mathbf{p}_C$ is an evolution path that is updated as
\begin{equation*}
  \mathbf{p}_C^{t+1} = (1-c_c) \mathbf{p}_C^{t} + \sqrt{\frac{c_c(2-c_c)}{\sum_{i=1}^\lambda \mathrm{w}_{i}^{2}}} \frac{\mathbf{m}^{t+1}-\mathbf{m}^t}{\sigma^{t}}.
\end{equation*}
Here $c_c$ is the learning rate for the evolution path update.
\end{description}
\end{enumerate}

The parameter adaptation in the CMA-ES is based on two principles. The first one is the maximum likelihood estimation (MLE). The update rules for $\mathbf{m}$ and the third term of the covariance matrix adaptation, called rank-$\mu$ update, can be interpreted as MLE. They are adapted so that it increases a weighted log-likelihood of previous samples, where points with higher fitness value have greater weights. The second one is the accumulation of successful steps. The step-size adaptation and the second term of the covariance matrix adaptation, called rank-one update, rely on the paths $\mathbf{p}_\sigma$ and $\mathbf{p}_C$. They are called evolution paths. Evolution paths contain information about the correlation between successive successful steps. Although evolution paths are reported to be unstable when $\lambda$ is large \cite{HansenEC2003,AkimotoGECCO2008}, they have a large effect on search speed and accuracy when $\lambda$ is small. 

In what follows, we investigate a simplified CMA-ES called rank-$\mu$ only CMA-ES in which the global step-size and evolution paths are removed. The resulting update rules reduce to 
\begin{align}
 \mathbf{m}^{t+1} &= \mathbf{m}^{t} + \eta_m \sum_{i=1}^\lambda \mathrm{w}_{R_i} (\mathbf{x}_{i} - \mathbf{m}^{t}) \label{eq:cma-m} \\
 \mathbf{C}^{t+1} &= \mathbf{C}^{t} + \eta_C \sum_{i=1}^\lambda \mathrm{w}_{R_i} \left((\mathbf{x}_{i} - \mathbf{m}^{t}) (\mathbf{x}_{i} - \mathbf{m}^{t})^\mathrm{T} - \mathbf{C}^{t}\right), \label{eq:cma-c}
\end{align}
where $\eta_m$ and $\eta_C$ are learning rate parameters.
\section{Natural Gradient Learning for Expected Fitness Maximization}
\label{sec:ngl}

In this section, we introduce natural gradient learning for expected fitness maximization. But, we start with the definition of statistical manifolds and the concept behind the natural gradient. Then we formulate an expected fitness maximization and the framework of natural gradient learning.

%\subsection{Brief Introduction to Information Geometry}
\paragraph{Statistical Manifold.} 
Information geometry \cite{Amari2007book} is the study of the natural differentiable geometric structure of manifolds of probability distributions. Consider a family $S$ of probability distributions on $\mathbb{R}^d$ parameterized using $n$ real-valued variables $\theta = [\theta_1 \dots \theta_n]$ so that
%\begin{equation}
$S = \{p_\theta = p(\mathbf{x}; \theta) \mid \theta \in \Theta\}$,
%\end{equation}
where $\Theta$ is a subset of $\mathbb{R}^n$ and the mapping $\theta \mapsto p_\theta$ is an injection. Such a set $S$ is called an $n$-dimensional statistical model on $\mathbb{R}^d$. The mapping $\varphi: S \to \mathbb{R}^n$ defined by $\varphi(p_\theta) = \theta$ is viewed as a coordinate system for $S$. With a Riemannian metric, termed Fisher metric, defined by the Fisher information matrix
\begin{equation}
	\mathbf{F}(\theta) = \int \frac{\partial \ln p(\mathbf{x}; \theta)}{\partial \theta}\left( \frac{\partial \ln p(\mathbf{x}; \theta)}{\partial \theta}\right)^\mathrm{T} p(\mathbf{x}; \theta)\ \mathrm{d}\mathbf{x},
\end{equation}
we can consider $S$ as a Riemannian manifold and then we call $S$ a statistical manifold.

It is possible to define an infinite number of Riemannian metrics on $S$. However, we find that there are properties that distinguish the Fisher metric from other metrics. One good property is that the Fisher metric is the only invariant metric under the choice of coordinate system \cite[Section~2.4]{Amari2007book}. The invariance is important in order to consider the intrinsic geometric structure of manifolds. The fact that the Fisher information matrix is the curvature of the KL-divergence \cite[Section~2.6]{Kullback1959book} is also a supportive property because the KL-divergence is commonly used to measure the difference between two probability distributions. Hence, the Fisher metric is considered as the most natural Riemannian metric on statistical manifolds.

\paragraph{Natural Gradient.} 
Consider $\rho$ as a function defined on a Riemannian manifold $S$ equipped with a Riemannian metric $\mathbf{G}$ with coordinate system $\varphi: p_{\theta} \mapsto \theta$. Let $\rho_{\varphi}(\theta) = \rho\left(\varphi^{-1}(\theta)\right)$. On the Riemannian manifold $S$, the steepest ascent direction of $\rho_{\varphi}$ is not usually given by the conventional gradient direction $\nabla \rho_{\varphi}(\theta)$. The natural gradient \cite{AmariNC1998}
\begin{equation}
 \tilde \nabla \rho_{\varphi}(\theta) = \mathbf{G}^{-1}(\theta) \nabla \rho_{\varphi}(\theta)
  \label{eq:ng}
\end{equation}
gives the steepest ascent direction of $\rho_{\varphi}$ on ($S$, $\mathbf{G}$) and it is invariant under the choice of coordinate system. %If the Riemannian metric is appropriate, it is expected that the natural gradient ascent leads to faster convergence than the vanilla gradient ascent. 
Natural gradient learning has been used as an efficient learning algorithm in several fields of machine learning \cite{AmariNC1998,AmariASSP1998,PetersNC2008}.

%\subsection{Natural Gradient Learning on Expected Fitness Maximization}
\paragraph{Expected Fitness.} 
Let $\pi(\mathbf{x}; \theta) = \pi\left(\mathbf{x}; \mathbf{m}(\theta), \mathbf{C}(\theta)\right)$ and $\Theta$ be a set of $\theta$ where $\mathbf{C}(\theta)$ is nonsingular. Then, the expected fitness with respect to $\pi(\mathbf{x}; \theta)$ is defined as
\begin{equation}
 J(\theta) = \mathbb{E}[f(\mathbf{x}); \theta] = \int f(\mathbf{x}) \pi(\mathbf{x}; \theta)\ \mathrm{d}\mathbf{x}.
\label{eq:ef}
\end{equation}
The function $J(\cdot)$ can be considered as a function on a statistical manifold. 

\paragraph{Natural Gradient Learning for Expected Fitness Maximization.} 
Since the metric $\mathbf{G}(\theta)$ on a statistical manifold is given by the Fisher information matrix $\mathbf{F}(\theta)$, the steepest ascent direction can be given by the natural gradient $\tilde \nabla J(\theta) = \mathbf{F}^{-1}(\theta) \nabla J(\theta)$. For the case of normal distributions, the $(i, j)^\text{th}$ element of the Fisher information matrix has a well-known explicit form \cite[p.~47 and Appendix~3C]{Kay1993book} 
\begin{equation}
 \mathbf{F}_{i,j}(\theta) = \frac{\partial \mathbf{m}^\mathrm{T}}{\partial \theta_i} \mathbf{C}^{-1} \frac{\partial \mathbf{m}}{\partial \theta_j} + \frac{1}{2} \text{tr} \left( \mathbf{C}^{-1} \frac{\partial \mathbf{C}}{\partial \theta_i} \mathbf{C}^{-1} \frac{\partial \mathbf{C}}{\partial \theta_j} \right).
\label{eq:fim-elem-normal}
\end{equation}
The gradient can be expressed as
\begin{align}
 \nabla J(\theta) = \nabla \int f(\mathbf{x}) \pi(\mathbf{x}; \theta)\ \mathrm{d}\mathbf{x} &= \int f(\mathbf{x}) \nabla \pi(\mathbf{x}; \theta)\ \mathrm{d}\mathbf{x} \notag\\
 &= \int f(\mathbf{x}) \pi(\mathbf{x}; \theta) \nabla \ln \pi(\mathbf{x}; \theta)\ \mathrm{d}\mathbf{x},
\label{eq:grad}
\end{align}
where the second equality holds under some regularity conditions which are derived from Lebesgue's dominated convergence theorem (see e.g.~\cite[Theorem~16.3]{Billingsley1995book}). Therefore, the natural gradient is expressed as
\begin{equation}
\tilde \nabla J(\theta) = \int f(\mathbf{x}) \mathbf{F}^{-1}(\theta) \nabla \ln \pi(\mathbf{x}; \theta) \pi(\mathbf{x}; \theta) \mathrm{d}\mathbf{x}.
\label{eq:ng-exact}
\end{equation}
Since the fitness function is unknown, so is the expected fitness and its natural gradient. We estimate the natural gradient by the Monte-Carlo approximation: 
\begin{equation}
\delta(\theta\mid\{\mathbf{x}_i\}) = \sum_{i=1}^\lambda \frac{f(\mathbf{x}_i)}{\lambda} \mathbf{F}^{-1}(\theta) \nabla \ln \pi(\mathbf{x}_i; \theta). 
\label{eq:ng-estim}
\end{equation}
Here, we can calculate the inverse of the Fisher information matrix \eqref{eq:fim-elem-normal} not necessarily analytically but numerically. Using the estimate $\delta(\theta\mid\{\mathbf{x}_i\})$ natural gradient learning for expected fitness maximization adjusts the parameter $\theta$ in the following rule: $\theta^{t+1} = \theta^{t} + \eta \delta(\theta^t\mid\{\mathbf{x}_i\})$.

\paragraph{Natural Evolution Strategies.} 
NESs adjust the parameters on the basis of the natural gradient on the expected fitness, but they non-linearly transform the fitness function. In the Monte-Carlo approximation of the natural gradient \eqref{eq:ng-estim}, NESs replace $f(\mathbf{x}_i)/\lambda$ with a ranking based weight $\mathrm{w}_{R_i}$. We call this transformation ranking based fitness shaping. The fitness shaping makes NESs enjoy the invariance property under order preserving, i.e.\ monotone, transformation of fitness function, as done in the CMA-ES.

% Section 4 %%%%%%%%%%%%%%%%%%%%%%%%%%%%%%%%%%%%%%%%%%%%%%%%%%%%%%%%%%%%%%%%%%%
\section{Analogy of the CMA-ES to Natural Gradient Learning}
\label{sec:natural}

This section discusses the analogy between the CMA-ES and natural gradient learning, which follows from the derivation of the explicit form of the natural gradient on the expected fitness. At the end of the section, we remark on some variants of the CMA-ES.

\subsection{General Form of the Natural Gradient}
\label{sec:natural:gf}

Let $\Theta$ be a set of parameters $\theta$ such that the normal distribution $\pi(\mathbf{x}; \theta)$ is nonsingular; i.e., the Fisher information matrix $\mathbf{F}(\theta)$ is nonsingular. We suppose that the parameter vector is divided into two parts $[\theta_m^\mathrm{T}, \theta_C^\mathrm{T}]^\mathrm{T}$, and
\begin{equation}
 \frac{\partial \mathbf{m}}{\partial \theta_C^\mathrm{T}} = \mathbf{0} \quad \text{and} \quad \frac{\partial \text{vech}(\mathbf{C})}{\partial \theta_m^\mathrm{T}}  = \mathbf{0}
\label{eq:asm}
\end{equation}
hold at $\theta \in \Theta$, where $vech$ denotes the half-vectorization operator that maps a $d$-dimensional square matrix to a $d(d+1)/2$-dimensional column vector that stacks columns starting at the diagonal elements of the matrix (see e.g., \cite[Chapter~16]{Harville2008book}). The assumption \eqref{eq:asm} is satisfied if $\mathbf{m}$ and $\mathbf{C}$ only depend on $\theta_m$ and $\theta_C$, respectively, which is satisfied in the cases that we treat in the later sections. Then, the Fisher information matrix has the block form $\mathbf{F}(\theta) = \text{diag}(\mathbf{F}_m(\theta), \mathbf{F}_{C}(\theta))$ and we have from \eqref{eq:ng-estim}
\begin{equation}
\delta(\theta\mid\{\mathbf{x}_i\}) = \sum_{i=1}^\lambda \frac{f(\mathbf{x}_i)}{\lambda} 
 \begin{bmatrix}
  \mathbf{F}_{m}^{-1}(\theta) \nabla_{\theta_m} \ln \pi(\mathbf{x}_i; \theta)\\
  \mathbf{F}_{C}^{-1}(\theta) \nabla_{\theta_C} \ln \pi(\mathbf{x}_i; \theta)\\
 \end{bmatrix}. 
\label{eq:ng-estim-block}
\end{equation}
Thus, we have the explicit form of the estimate of the natural gradient at $\theta$ if we can analytically evaluate each block of the right-hand side of \eqref{eq:ng-estim-block}. However, it is not trivial to calculate the inverse of the Fisher information matrix and express it in terms of $\mathbf{m}$ and $\mathbf{C}$.

The following theorem shows that we can directly obtain the product of the inverse of the Fisher information matrix and the gradient of the log-likelihood without inversion of the Fisher information matrix. 

% Begin Theorem %%%%%%%%%%%%%%%%%%%%%%%%%%%%%%%%%%%%%%%%%%%%%%%%%%%%%%%%%%%%%%%
\begin{thm}
\label{thm:1}
Suppose $\theta_m$ and $\theta_C$ are $d$- and $d(d+1)/2$-dimensional column vectors, respectively. Then $\partial \mathbf{m} / \partial \theta_m^\mathrm{T}$ and $\partial \text{vech}(\mathbf{C}) / \partial \theta_C^\mathrm{T}$ are invertible at $\theta \in \Theta$, and 
\begin{align}
 \mathbf{F}_m^{-1}(\theta) \nabla_{\theta_m} \ln \pi(\mathbf{x} \mid \theta) &= \left(\frac{\partial \mathbf{m}}{\partial \theta_m^\mathrm{T}} \right)^{-1} (\mathbf{x} - \mathbf{m}) \label{eq:thm-ng-mean}\\
 \mathbf{F}_C^{-1}(\theta) \nabla_{\theta_C} \ln \pi(\mathbf{x} \mid \theta) &= \left(\frac{\partial \text{vech}(\mathbf{C})}{\partial \theta_C^\mathrm{T}} \right)^{-1} \text{vech}\left((\mathbf{x} - \mathbf{m})(\mathbf{x} - \mathbf{m})^\mathrm{T} - \mathbf{C} \right). \label{eq:thm-ng-full}
\end{align}
\end{thm}
% End Theorem %%%%%%%%%%%%%%%%%%%%%%%%%%%%%%%%%%%%%%%%%%%%%%%%%%%%%%%%%%%%%%%%%

Theorem~\ref{thm:1} shows that if the derivatives of the mean vector and the covariance matrix with respect to $\theta_m$ and $\theta_C$ have simple forms and their inverse matrices can be easily expressed in terms of $\mathbf{m}$ and $\mathbf{C}$, then we can obtain the form of the natural gradient \eqref{eq:ng-exact} and the estimate $\delta(\theta\mid\{\mathbf{x}_i\})$ analytically by using \eqref{eq:thm-ng-mean} and \eqref{eq:thm-ng-full}. In most cases, the additional inversion is easier to perform than the inversion of the Fisher information matrix.

It is worth mentioning that the natural gradient can be also derived by the way taken by Glasmachers~et~al.~\cite{GlasmachersGECCO2010}. To avoid the computation of the Fisher information matrix, they introduce a local coordinate on $S$ where the Fisher information matrix is identical to the unit matrix. They show the statement of the natural gradient under exponential parameterization described in Section~\ref{sec:natural:remarks}.

% Begin Proof %%%%%%%%%%%%%%%%%%%%%%%%%%%%%%%%%%%%%%%%%%%%%%%%%%%%%%%%%%%%%%%%%
\begin{proof}
First, we derive the inverse matrix of each block of the Fisher information matrix. From \eqref{eq:fim-elem-normal} and assumption \eqref{eq:asm} we have the block of the Fisher information matrix corresponding to $\theta_{m}$
\begin{equation}
 \mathbf{F}_m = \left( \frac{\partial \mathbf{m}}{\partial \theta_{m}^\mathrm{T}} \right)^\mathrm{T} \mathbf{C}^{-1} \left( \frac{\partial \mathbf{m}}{\partial \theta_{m}^\mathrm{T}} \right).
\label{eq:fim-m-mat}
\end{equation}
Since $\partial \mathbf{m}/\partial \theta_{m}^\mathrm{T}$ is a $d$-dimensional square matrix, it must be invertible if $\mathbf{F}_m$ is invertible. Since $\mathbf{F}$ is nonsingular at $\theta \in \Theta$, $\mathbf{F}_m$ is invertible. Thus, $\partial \mathbf{m}/\partial \theta_{m}^\mathrm{T}$ is invertible. Then, the inverse matrix of $\mathbf{F}_m$ is expressed as
\begin{equation}
 \mathbf{F}_m^{-1} = \left(\frac{\partial \mathbf{m}}{\partial \theta_{m}^\mathrm{T}}\right)^{-1} \mathbf{C} \biggl[\biggl(\frac{\partial \mathbf{m}}{\partial \theta_{m}^\mathrm{T}}\biggl)^{-1}\biggr]^\mathrm{T}.
\label{eq:fim-mean-inv}
\end{equation}
From \eqref{eq:fim-elem-normal}, assumption \eqref{eq:asm}, and the formula of matrix differentiation
(see e.g., \cite[Chapter~15]{Harville2008book})
\begin{equation}
 \frac{\partial \mathbf{C}^{-1}}{\partial \theta_i} = - \mathbf{C}^{-1} \frac{\partial \mathbf{C}}{\partial \theta_i} \mathbf{C}^{-1}, \label{eq:deriv-of-inv}
\end{equation}
we have the $(i, j)^\text{th}$ element of the block of the Fisher information matrix corresponding to $\theta_{C}$ as
\begin{equation*}
\begin{split}
 (\mathbf{F}_C)_{i,j} &= \frac{1}{2} \text{tr} \left( \mathbf{C}^{-1} \frac{\partial \mathbf{C}}{\partial \theta_{C,i}} \mathbf{C}^{-1} \frac{\partial \mathbf{C}}{\partial \theta_{C,j}} \right) 
  = - \frac{1}{2} \text{tr} \left( \frac{\partial \mathbf{C}^{-1}}{\partial \theta_{C,i}} \frac{\partial \mathbf{C}}{\partial \theta_{C,j}} \right) \\
  &= - \frac{1}{2} \text{vech}\left( 2\frac{\partial \mathbf{C}^{-1}}{\partial \theta_{C,i}} - \text{diag}\left( \frac{\partial \mathbf{C}^{-1}}{\partial \theta_{C,i}} \right)\right)^\mathrm{T} \text{vech}\left( \frac{\partial \mathbf{C}}{\partial \theta_{C,j}}\right) \\
  &= - \frac{1}{2} \left(\frac{\partial \text{vech}( 2\mathbf{C}^{-1} - \text{diag}( \mathbf{C}^{-1} ))}{\partial \theta_{C,i}}\right)^\mathrm{T} \frac{\partial \text{vech}( \mathbf{C})}{\partial \theta_{C,j}}, 
\end{split}
\end{equation*}
where $\text{diag}(\mathbf{C})$ represents a diagonal matrix whose diagonal elements equal the diagonal elements of $\mathbf{C}$. Then, we have the matrix form
\begin{equation}
 \mathbf{F}_C = - \frac{1}{2} \left(\frac{\partial \text{vech}( 2\mathbf{C}^{-1} - \text{diag}( \mathbf{C}^{-1} ))}{\partial \theta_{C}^\mathrm{T}}\right)^\mathrm{T} \frac{\partial \text{vech}( \mathbf{C})}{\partial \theta_{C}^\mathrm{T}}. 
\end{equation}
Since both $\partial \text{vech}( 2\mathbf{C}^{-1} - \text{diag}( \mathbf{C}^{-1}))/ \partial \theta_C^\mathrm{T}$ and $\partial \text{vech}(\mathbf{C}) / \partial \theta_C^\mathrm{T}$ are square matrices of dimension $d(d+1)/2$, they must be invertible if $\mathbf{F}_C$ is invertible. By the assumption \eqref{eq:asm}, $\mathbf{F}$ is invertible for $\theta \in \Theta$, and hence, so is $\mathbf{F}_C$. Thus, $\partial \text{vech}( 2\mathbf{C}^{-1} - \text{diag}( \mathbf{C}^{-1}))/ \partial \theta_C^\mathrm{T}$ and $\partial \text{vech}(\mathbf{C}) / \partial \theta_C^\mathrm{T}$ are invertible and the inverse of $\mathbf{F}_{C}$ is expressed as
\begin{equation}
  \mathbf{F}_C^{-1} = - 2 \left( \frac{\partial \text{vech}(\mathbf{C})}{\partial \theta_C^\mathrm{T}}\right)^{-1} \biggl[\left(\frac{\partial \text{vech}( 2\mathbf{C}^{-1} - \text{diag}( \mathbf{C}^{-1}))}{\partial \theta_C^\mathrm{T}}\right)^{-1}\biggr]^\mathrm{T}.
\label{eq:fim-inv-full}
\end{equation}

Next, we derive each block of the gradient of the log-likelihood $\ln \pi(\mathbf{x}; \theta)$. The log-likelihood function for the normal distribution is written as
\begin{equation}
 \ln \pi(\mathbf{x}; \theta) = - \frac{d \ln 2\pi}{2} - \frac{\ln \det{\mathbf{C}}}{2} - \frac{\text{tr}(\mathbf{C}^{-1}(\mathbf{x} - \mathbf{m})(\mathbf{x} - \mathbf{m})^\mathrm{T})}{2}.
\label{eq:loglike}
\end{equation}
Then, in light of formula \eqref{eq:deriv-of-inv} and another formula of matrix differentiation (see e.g., \cite[Chapter~15]{Harville2008book})
\begin{equation*}
 \frac{\partial \ln \det{\mathbf{C}}}{\partial \theta_i} = \text{tr}\left(\mathbf{C}^{-1} \frac{\partial \mathbf{C}}{\partial \theta_i}\right),
\end{equation*}
the partial derivative of \eqref{eq:loglike} with respect to $\theta_i$ can be written in the form
\begin{equation}
 \frac{\partial \ln \pi(\mathbf{x}; \theta)}{\partial \theta_i} = - \frac{1}{2}\text{tr}\left(\frac{\partial \mathbf{C}^{-1}}{\partial \theta_i} \left((\mathbf{x} - \mathbf{m})(\mathbf{x} - \mathbf{m})^\mathrm{T} - \mathbf{C}\right)\right) + \frac{\partial \mathbf{m}^\mathrm{T}}{\partial \theta_i} \mathbf{C}^{-1} (\mathbf{x} - \mathbf{m}).
\label{eq:deriv-loglike}
\end{equation}
According to assumption \eqref{eq:asm}, we have 
\begin{equation}
 \nabla_{\theta_m} \ln \pi(\mathbf{x}; \theta) = \frac{\partial \ln \pi(\mathbf{x}; \theta)}{\partial \theta_m^\mathrm{T}} = \left(\frac{\partial \mathbf{m}}{\partial \theta_m^\mathrm{T}}\right)^\mathrm{T} \mathbf{C}^{-1} (\mathbf{x} - \mathbf{m}).
\label{eq:grad-loglike-m}
\end{equation}
By rewriting the first term of \eqref{eq:deriv-loglike} as
\begin{multline*}
- \frac{1}{2}\text{tr}\left(\frac{\partial \mathbf{C}^{-1}}{\partial \theta_i} \left((\mathbf{x} - \mathbf{m})(\mathbf{x} - \mathbf{m})^\mathrm{T} - \mathbf{C}\right)\right) \\
 = - \frac{1}{2} \left(\frac{\partial \text{vech}( 2\mathbf{C}^{-1} - \text{diag}( \mathbf{C}^{-1} ))}{\partial \theta_{C,i}}\right)^\mathrm{T} \text{vech}\left((\mathbf{x} - \mathbf{m})(\mathbf{x} - \mathbf{m})^\mathrm{T} - \mathbf{C}\right),
\end{multline*}
we have the block of the gradient corresponding to $\theta_C$ as follows
\begin{multline}
 \nabla_{\theta_C} \ln \pi(\mathbf{x}; \theta) = \frac{\partial \ln \pi(\mathbf{x}; \theta)}{\partial \theta_C^\mathrm{T}} = - \frac{1}{2} \left(\frac{\partial \text{vech}( 2\mathbf{C}^{-1} - \text{diag}( \mathbf{C}^{-1} ))}{\partial \theta_{C}^\mathrm{T}}\right)^\mathrm{T} \\ \cdot \text{vech}\left((\mathbf{x} - \mathbf{m})(\mathbf{x} - \mathbf{m})^\mathrm{T} - \mathbf{C}\right).
\label{eq:grad-loglike-c}
\end{multline}

Taking the product of \eqref{eq:fim-mean-inv} and \eqref{eq:grad-loglike-m} and the product of \eqref{eq:fim-inv-full} and \eqref{eq:grad-loglike-c},
we have finally \eqref{eq:thm-ng-mean} and \eqref{eq:thm-ng-full}. This completes the proof.\qed
\end{proof}
% End Proof %%%%%%%%%%%%%%%%%%%%%%%%%%%%%%%%%%%%%%%%%%%%%%%%%%%%%%%%%%%%%%%%%%%

\subsection{Theoretical Foundation for the Parameter Update in the CMA-ES}
\label{sec:natural:cma}

Theorem~\ref{thm:1} is useful to derive the explicit form of the natural gradient learning algorithm under some parameterization. Consider one of the simplest parameterization: $\mathbf{m}(\theta) = \theta_m$ and $\text{vech}(\mathbf{C}(\theta)) = \theta_C$. Since $\partial \mathbf{m}/ \partial \theta_m^\mathrm{T} = \mathbf{I}$ and $\partial \text{vech}(\mathbf{C})/ \partial \theta_C^\mathrm{T} = \mathbf{I}$, from \eqref{eq:ng-estim-block}, \eqref{eq:thm-ng-mean}, and \eqref{eq:thm-ng-full}, we have the update rules for natural gradient learning
\begin{equation}
\theta^{t+1} = \theta^{t} + \eta \sum_{i=1}^\lambda \frac{f(\mathbf{x}_i)}{\lambda} 
\begin{bmatrix}
 \mathbf{x}_i - \mathbf{m}(\theta^t)\\
 \text{vech}\left((\mathbf{x}_i - \mathbf{m}(\theta^t))(\mathbf{x}_i - \mathbf{m}(\theta^t))^\mathrm{T} - \mathbf{C}(\theta^t)\right)
\end{bmatrix}.
\label{eq:ngl-cma}
\end{equation}
Let $\mathbf{m}^{t} = \mathbf{m}(\theta^t)$ and $\mathbf{C}^{t} = \mathbf{C}(\theta^t)$. Separating \eqref{eq:ngl-cma} into an $\mathbf{m}$-part and $\mathbf{C}$-part, we have
\begin{align}
\mathbf{m}^{t+1} &= \mathbf{m}^{t} + \eta \sum_{i=1}^\lambda \frac{f(\mathbf{x}_i)}{\lambda} (\mathbf{x}_i - \mathbf{m}^t)
\label{eq:ngl-cma-m}
\\
\mathbf{C}^{t+1} &= \mathbf{C}^{t} + \eta \sum_{i=1}^\lambda \frac{f(\mathbf{x}_i)}{\lambda} \left((\mathbf{x}_i - \mathbf{m}^t)(\mathbf{x}_i - \mathbf{m}^t)^\mathrm{T} - \mathbf{C}^t\right).
\label{eq:ngl-cma-c}
\end{align}

We notice that the update rules \eqref{eq:cma-m} and \eqref{eq:cma-c} in the CMA-ES are the same as \eqref{eq:ngl-cma-m} and \eqref{eq:ngl-cma-c} derived from natural gradient learning, except that the CMA-ES uses ranking-based weights $\mathrm{w}_{R_i}$ instead of raw fitness values $f(\mathbf{x}_i)/\lambda$ and employs different learning rates for $\mathbf{m}$ and $\mathbf{C}$. In other words, when using a common value $\eta_m = \eta_C = \eta$ and assigning $\mathrm{w}_{R_i} = f(\mathbf{x}_i) / \lambda$ for every iteration, the rank-$\mu$ only CMA-ES updates the distribution parameters along the sampled natural gradient of the expected fitness. 

The coefficients $f(\mathbf{x}_i) / \lambda$ in natural gradient learning approximately sum up to $J(\theta)$, because $\sum_{i=1}^{\lambda}f(\mathbf{x}_i) / \lambda$ is a Monte-Carlo estimate of the expected fitness \eqref{eq:ef}, and they increase as the expected fitness increases. On the contrary, the weights $\mathrm{w}_{i}$ in the CMA-ES are fixed and sum up to one. Therefore, with the fixed learning rates, the adjustment for the parameters in the CMA-ES is approximately $1/J(\theta)$ times as large as that in \eqref{eq:ngl-cma-m} and \eqref{eq:ngl-cma-c}. Providing that $J(\theta)$ is positive, this corresponds to the relation between $\tilde \nabla J(\theta)$ and $\tilde \nabla \ln J(\theta) = \tilde \nabla J(\theta) / J(\theta)$. By replacing $\tilde \nabla J(\theta)$ and $J(\theta)$ with their Monte-Carlo estimates $\delta(\theta\mid\{\mathbf{x}_i\})$ and $\hat J(\theta\mid\{\mathbf{x}_i\}) = \sum_{i=1}^\lambda f(\mathbf{x}_i)/\lambda$, we have a sampled natural gradient of the log of expected fitness:
\begin{equation}
\delta_{\ln J}(\theta\mid\{\mathbf{x}_i\}) = \sum_{i=1}^\lambda \frac{f(\mathbf{x}_i)}{\sum_{j=1}^\lambda f(\mathbf{x}_j)} \mathbf{F}^{-1}(\theta) \nabla \ln \pi(\mathbf{x}_i; \theta). 
\label{eq:log-ng-estim}
\end{equation}
Then, we obtain the update rules for the $\mathbf{m}$ and $\mathbf{C}$-parts by replacing $f(\mathbf{x}_i)/\lambda$ in \eqref{eq:ngl-cma-m} and \eqref{eq:ngl-cma-c} with $f(\mathbf{x}_i) / \sum_{j=1}^\lambda f(\mathbf{x}_j)$. We notice a closer relation between the CMA-ES and the natural gradient of the log of expected fitness: not only are the forms of their learning rules the same, but the coefficients in natural gradient learning using \eqref{eq:log-ng-estim} also share properties with the commonly-used weight setting in the CMA-ES. 

However, this algorithm is not invariant under monotone transformation of fitness function, whereas the CMA-ES is invariant under such transformation and the invariance is an important property of the CMA-ES. More study about the coefficients is an important future work.
% Weights fixed during learning can be interpreted as the result of the estimate of the natural gradient of an expected fitness, where the definition of the fitness function $f$, i.e., the choice of a strictly increasing function $g$, changes for every iteration.

In short, this result provides a theoretical justification for the parameter update in the rank-$\mu$ only CMA-ES. Since the natural gradient points to the steepest ascent direction of a function defined on a Riemannian manifold, the CMA-ES turns out to be based on a steepest ascent method with sampled natural gradient of (the log of) the expected fitness on the parameter space, which is a well-principled approach. 

\subsection{Remarks}
\label{sec:natural:remarks}

There are some remarks that can be made on the results.

\paragraph{CMA-ES and NES.} 

Now that we have found the CMA-ES is based on the sampled natural gradient on the expected fitness, it is clear that the CMA-ES can be considered a variant of NESs. With the same fitness shaping (mapping raw fitness values to ranking-based weights), the rank-$\mu$ only CMA-ES can be described in the framework of NESs. The original NES \cite{WierstraCEC2008} and efficient NES (eNES) \cite{SunGECCO2009,SunICML2009} use Cholesky parameterization: $\text{vech}(\mathbf{A}) = \theta_C$, where $\mathbf{A}$ is the (lower triangular) Cholesky factor satisfying $\mathbf{C} = \mathbf{A}\mathbf{A}^\mathrm{T}$. Exponential NES (xNES) \cite{GlasmachersGECCO2010} employs exponential parameterization $\text{vech}(\mathbf{B}) = \theta_C$, where $\mathbf{C} = \exp(\mathbf{B})$, and the CMA-ES parameterizes the distribution by $\text{vech}(\mathbf{C}) = \theta_C$. Although the natural gradient itself is invariant under the choice of coordinate system, a finite step along the natural gradient leads to a slightly different learning rule under nonlinear transformation of the coordinate system as done in eNES, xNES, and the CMA-ES. 

\paragraph{Restricted Coordinate System}

For some restricted covariance matrix cases, we can attain the corresponding form of the natural gradient in the same manner as in the proof of Theorem~\ref{thm:1}. For instance, if $\theta_C$ is a scalar and $\mathbf{C}(\theta) = \sigma(\theta_C) \mathbf{C}_0$, where $\sigma$ is a function and $\sigma(\theta_C) > 0$ for $\theta \in \Theta$, and $\mathbf{C}_0$ is fixed, we have
\begin{equation*}
 \mathbf{F}_C^{-1}(\theta) \nabla_{\theta_C} \ln \pi(\mathbf{x} \mid \theta) = \left( \frac{\partial \sigma}{\partial \theta_{C}} \right)^{-1} \biggl(\frac{(\mathbf{x} - \mathbf{m})^\mathrm{T} \mathbf{C}_0^{-1} (\mathbf{x} - \mathbf{m})}{d} - \sigma \biggr).
\label{eq:thm-ng-single}
\end{equation*}
For instance, if $\theta_C$ is a $d$-dimensional column vector and $\mathbf{C}(\theta)$ is a diagonal matrix whose $i^\text{th}$ diagonal element is $\sigma_i(\theta)$, where $\sigma_i$ are functions such that $\sigma_i(\theta) > 0$ for $\theta \in \Theta$, we have
 \begin{equation*}
  \mathbf{F}_C^{-1} \nabla_{\theta_C} \ln \pi(\mathbf{x} \mid \theta) = 
   \biggl[\biggl( \frac{\partial [\sigma_1, \dots, \sigma_d]}{\partial \theta_{C}} \biggr)^{-1}\biggr]^\mathrm{T} \left[(\mathbf{x} - \mathbf{m})_1^2 - \sigma_1, \dots, (\mathbf{x} - \mathbf{m})_d^2 - \sigma_d\right]^\mathrm{T}.
   \label{eq:thm-ng-diag}
 \end{equation*}

\paragraph{sep-CMA-ES and Restricted Coordinate System.}

Ros and Hansen \cite{RosPPSN2008} proposed a variant of the CMA-ES, named sep-CMA-ES, in which the covariance matrix is constrained to be diagonal. The sep-CMA-ES without the rank-one update \cite{HansenEC2001} updates the diagonal elements $\sigma_i$ of the covariance matrix $\mathbf{C} = \text{diag}(\sigma_1, \dots, \sigma_d)$ as follows:
\begin{equation*}
 \sigma_i^{t+1} = \sigma_i^{t} + \eta_C \sum_{i=1}^\lambda \mathrm{w}_{R_i} \left((\mathbf{x} - \mathbf{m})_i^2 - \sigma_i\right).
\end{equation*}
This is the same as the covariance update rule derived from natural gradient learning when using a diagonal parameterization: $\mathbf{C}(\theta) = \text{diag}(\theta_{C,1}, \dots, \theta_{C,d})$.

\paragraph{Active-CMA-ES and Fitness Baseline.}

Consider the following equalities
\begin{equation*}
\begin{split}
 \mathbb{E}[(f(\mathbf{x}) - b) \nabla \ln \pi(\mathbf{x}; \theta)] &= \nabla \mathbb{E}[f(\mathbf{x}) - b] = \nabla \mathbb{E}[f(\mathbf{x})] - \nabla b \\
 &= \nabla \mathbb{E}[f(\mathbf{x})] = \mathbb{E}[f(\mathbf{x}) \nabla \ln \pi(\mathbf{x}; \theta)].
\end{split}
\end{equation*}
Thus, subtraction of $b$ from the fitness does not affect the expectation of the gradient estimation but does affect the variance of the estimation. This fact is used to reduce the variance of Monte-Carlo estimates and $b$ is referred to as a baseline (see e.g., \cite{Evans2000book,PetersNC2008,SunICML2009}). The natural gradient view and this fact clarify the relation between the CMA-ES and active-CMA-ES \cite{JastrebskiCEC2006}. Active-CMA-ES was proposed to reduce covariance adaptation time by reducing actively the elements of the covariance matrix corresponding to unsuccessful search directions and is implemented by using weights $\mathrm{w}_{R_i}$ that are possibly negative and sum up to zero, whereas they are nonnegative and sum up to one in the CMA-ES. When the weights in active-CMA-ES are equal to the weights in the CMA-ES minus some value, active-CMA-ES and CMA-ES estimate the same natural gradient with and without a baseline.

% Section 5 %%%%%%%%%%%%%%%%%%%%%%%%%%%%%%%%%%%%%%%%%%%%%%%%%%%%%%%%%%%%%%%%%%%
\section{Correspondence to the Generalized Expectation Maximization}
\label{sec:improvement}

In this section, we discuss the learning rates for natural gradient learning for expected fitness maximization. We derive the range of learning rates that ensure monotonic improvement in the expected fitness if the exact natural gradient is given. Then, we validate the setting of learning rates used in the CMA-ES. Finally, we discuss the relation to the fitness expectation maximization algorithm \cite{WierstraPPSN2008}, which is an EM-inspired algorithm for continuous optimization, and provide the information geometric interpretation of the CMA-ES.

\subsection{Monotone Improvement in the Expected Fitness}

The learning rates in the CMA-ES are usually fixed during learning. They are small positive constants when the sample size $\lambda$ is small, and reach values up to one when the sample size is large. In addition, they are different for the mean vector and for the covariance matrix. Considering the analogy to natural gradient learning, such a setting of learning rates is exceptional since the optimal step-size (learning rate) generally varies with the function and the position, and different learning rates make the adjustment vector stray from the steepest gradient.

To confirm the validity of such setting for the learning rates, we derive the range of learning rates that guarantee monotonic increase in the expected fitness. Suppose that $f(\mathbf{x})$ is positive, which holds at least if one defines the fitness as $\exp(f(\mathbf{x}))$ instead of $f(\mathbf{x})$. Then $J(\theta) > 0$ holds and we can view $q(\mathbf{x}; \theta) = f(\mathbf{x})\pi(\mathbf{x}; \theta)/J(\theta)$ as a probability density function on $\mathbb{R}^{d}$ because $q(\mathbf{x}; \theta) > 0$ and $\int q(\mathbf{x}; \theta) d\mathbf{x} = 1$. To show that a step-by-step improvement in the expected fitness is guaranteed, we consider the following equality:
\begin{align}
 \ln \frac{J(\theta')}{J(\theta)} 
 &= \ln \frac{J(\theta')f(\mathbf{x}) \pi(\mathbf{x}; \theta) }{J(\theta)f(\mathbf{x}) \pi(\mathbf{x}; \theta')} + \ln\frac{\pi(\mathbf{x}; \theta')}{\pi(\mathbf{x}; \theta)} = \ln \frac{q(\mathbf{x}; \theta)}{q(\mathbf{x}; \theta')} + \ln \frac{\pi(\mathbf{x}; \theta')}{\pi(\mathbf{x}; \theta)}\notag\\
 &= \int q(\mathbf{x}; \theta) \left( \ln \frac{q(\mathbf{x}; \theta)}{q(\mathbf{x}; \theta')} + \ln \frac{\pi(\mathbf{x}; \theta')}{\pi(\mathbf{x}; \theta)}\right)\ \mathrm{d}\mathbf{x}\notag\\
 &= D_\mathrm{KL}\left(q(\mathbf{x}; \theta) \parallel q(\mathbf{x}; \theta')\right) + Q(\theta, \theta') - Q(\theta, \theta)\label{eq:improvement}
\end{align}
where $Q(\theta, \theta')$ denotes the negative cross entropy $-H\left(q(\mathbf{x}; \theta), \pi(\mathbf{x}; \theta')\right)$ of $q(\mathbf{x}; \theta)$ and $\pi(\mathbf{x}; \theta')$ defined by
\begin{equation*}
 Q(\theta, \theta') = - H\left(q(\mathbf{x}; \theta\right), \pi(\mathbf{x}; \theta')) = \int q(\mathbf{x}; \theta) \ln \pi(\mathbf{x}; \theta')\ \mathrm{d}\mathbf{x},
\end{equation*}
and $D_\mathrm{KL}(p_1 \parallel p_2)$ represents the Kullback-Leibler (KL) divergence of $p_2$ from $p_1$, defined by $D_\mathrm{KL}(p_1 \parallel p_2) = H(p_1, p_2) - H(p_1)$. Here $H(p_1)$ denotes the entropy of $p_1$. Since KL divergence is always non-negative, we have the following inequality
\begin{equation}
 \ln J(\theta') - \ln J(\theta) \geq Q(\theta, \theta') - Q(\theta, \theta)
\label{eq:lower-bound}
\end{equation}
with equality holding if and only if $\theta = \theta'$. Thus, if we can choose $\theta'$ repeatedly to satisfy $Q(\theta, \theta') \geq Q(\theta, \theta)$, then step-by-step progress is guaranteed from \eqref{eq:lower-bound}. 

If the natural gradient is estimated sufficiently well, an infinitesimal step in the direction leads to an increase in expected fitness. The following theorem shows how long a step we can take along the exact natural gradient so as to guarantee improvement in expected fitness.

% Begin Theorem %%%%%%%%%%%%%%%%%%%%%%%%%%%%%%%%%%%%%%%%%%%%%%%%%%%%%%%%%%%%%%%
\begin{thm}
\label{thm:2}
Assume that $J(\theta)$ is differentiable. For $\theta \in \Theta$, suppose $\mathbf{m}(\theta) = \theta_{m}$ and $\text{vech}(\mathbf{C}(\theta)) = \theta_{C}$, and let 
\begin{equation*}
 \theta'(\eta_m, \eta_C) = 
  \begin{bmatrix}
   \theta_m + \eta_m \tilde\nabla_{\theta_m} J(\theta)\\
   \theta_C + \eta_C \tilde\nabla_{\theta_C} J(\theta)
  \end{bmatrix}.
%\label{eq:thm2}
\end{equation*}
If $\tilde \nabla_{\theta_{C}} J(\theta) \neq \mathbf{0}$, then the mapping $\eta_C \mapsto Q(\theta, \theta'(0, \eta_c))$ is strictly increasing in $\eta_c \in (0, 1/J(\theta))$ and has a local maximum point at $\eta_c = 1/J(\theta)$. Moreover, if $\tilde \nabla_{\theta_{m}} J(\theta) \neq \mathbf{0}$, then for any $\eta_C \in [0, 1/J(\theta)]$ the map $\eta_m \mapsto Q(\theta, \theta'(\eta_m, \eta_C))$ is strictly increasing in $\eta_m \in (0, 1/J(\theta))$ and has a local maximum point at $\eta_m = 1/J(\theta)$.
\end{thm}
% End Theorem %%%%%%%%%%%%%%%%%%%%%%%%%%%%%%%%%%%%%%%%%%%%%%%%%%%%%%%%%%%%%%%%%

Note that Theorem~\ref{thm:2} does not necessarily hold under other types of parameterization such as Cholesky parameterization or exponential parameterization. This is because they lead to different trajectories, although these are considered as discretizations of the same associated ordinary differential equation. Additionally, note that $\eta_m = \eta_C = 1/J(\theta)$ gives a local maximum point of $Q(\theta, \theta'(\eta_m, \eta_C))$ in $\eta_{m}$ and $\eta_{C}$, but $Q(\theta, \bar\theta)$ itself does not have a local maximum point at $\bar\theta = \theta'(1/J(\theta), 1/J(\theta))$.

% Begin Proof %%%%%%%%%%%%%%%%%%%%%%%%%%%%%%%%%%%%%%%%%%%%%%%%%%%%%%%%%%%%%%%%%
\begin{proof}
Let $\mathbf{m}(\theta)$ and $\mathbf{C}(\theta)$ be denoted by $\mathbf{m}$ and $\mathbf{C}$ respectively, and $\mathbf{m}(\theta'(\eta_m, \eta_C))$ and $\mathbf{C}(\theta'(\eta_m, \eta_C))$ be denoted by $\mathbf{m}_{\eta_m}$ and $\mathbf{C}_{\eta_C}$ respectively. First, we prove the first half of the theorem. The derivative of $Q(\theta, \theta'(0, \eta_C))$ with respect to $\eta_C$ is
\begin{equation}
 \frac{\partial Q(\theta, \theta'(0, \eta_C))}{\partial \eta_C} = \frac{\tilde\nabla_{\theta_C} J(\theta)^\mathrm{T}}{J(\theta)} \int f(\mathbf{x}) \pi(\mathbf{x}; \theta) \nabla_{\theta_C} \ln \pi(\mathbf{x}; \theta'(0, \eta_C)) d\mathbf{x}.
\label{eq:deriv-q}
\end{equation} 
Since $\mathbf{m}_{0} = \mathbf{m}$ and $\text{vech}(\mathbf{C}_{\eta_C}) = \theta_C + \eta_C \tilde\nabla_{\theta_{C}} J(\theta) = \text{vech}(\mathbf{C}) + \eta_C \tilde\nabla_{\theta_{C}} J(\theta)$, by taking \eqref{eq:thm-ng-full} into account we have
\begin{equation*}
\begin{split}
\nabla_{\theta_C} \ln \pi(\mathbf{x} \mid \theta'(0, \eta_C)) &= \mathbf{F}_C(\theta'(0, \eta_C)) \mathbf{F}_C^{-1}(\theta'(0, \eta_C)) \nabla_{\theta_C} \ln \pi(\mathbf{x}; \theta'(0, \eta_C)) \\
&= \mathbf{F}_C(\theta'(0, \eta_C)) \text{vech}( (\mathbf{x} - \mathbf{m}) (\mathbf{x} - \mathbf{m})^\mathrm{T} - \mathbf{C}_{\eta_C})\\
&= \mathbf{F}_C(\theta'(0, \eta_C)) (\text{vech}( (\mathbf{x} - \mathbf{m}) (\mathbf{x} - \mathbf{m})^\mathrm{T} - \mathbf{C}) - \eta_C \tilde\nabla_{\theta_{C}} J(\theta))\\
&= \mathbf{F}_C(\theta'(0, \eta_C)) (\mathbf{F}_C(\theta) \nabla_{\theta_C} \ln \pi(\mathbf{x}; \theta) - \eta_C \tilde\nabla_{\theta_{C}} J(\theta)).
\end{split}
\end{equation*}
Since $\mathbb{E}[f(\mathbf{x})\mathbf{F}_C(\theta) \nabla_{\theta_C} \ln \pi(\mathbf{x}; \theta)] = \mathbf{F}_C(\theta) \nabla_{\theta_C} J(\theta) = \tilde \nabla_{\theta_C} J(\theta)$, where the expectation is taken under $\pi(\mathbf{x}; \theta)$, the derivative \eqref{eq:deriv-q} reduces to
\begin{equation}
 \frac{\partial Q(\theta, \theta'(0, \eta_C))}{\partial \eta_C} = \left(\frac{1}{ J(\theta)} - \eta_C\right) \tilde\nabla_{\theta_{C}} J(\theta)^\mathrm{T} \mathbf{F}_C(\theta'(0, \eta_C)) \tilde \nabla_{\theta_C} J(\theta).
\label{eq:deriv-q-2}
\end{equation}
Here, for $\eta_C \in [0, 1/J(\theta)]$,
\begin{equation*}
 \mathbf{C}_{\eta_C} = (1 - \eta_C J(\theta) )\mathbf{C} + \eta_C \mathbb{E}[f(\mathbf{x})(\mathbf{x} - \mathbf{m})(\mathbf{x} - \mathbf{m})^\mathrm{T}]
\end{equation*}
is positive definite because $(1 - \eta_C J(\theta) )\mathbf{C}$ is non-negative definite and $f(\mathbf{x}) > 0$ means $\mathbb{E}[f(\mathbf{x})(\mathbf{x} - \mathbf{m})(\mathbf{x} - \mathbf{m})^\mathrm{T}]$ is positive definite, and the sum of non-negative and positive definite matrices gives another positive definite matrix. From the continuity of the positivity, $\mathbf{C}_{\eta_C}$ is positive for $\eta_C \in [0, 1/J(\theta) + \epsilon)$ for small $\epsilon$. Hence, the Fisher information matrix $\mathbf{F}_C(\theta'(0, \eta_C))$ is also positive definite for $\eta_C \in [0, 1/J(\theta) + \epsilon)$. Thus, the right-hand side of equation \eqref{eq:deriv-q-2} is positive if $\eta_C \in [0, 1/J(\theta))$, zero if $\eta_C = 1/J(\theta)$, negative if $\eta_C \in (1/J(\theta), 1/J(\theta) + \epsilon)$. Consequently, we find that $Q(\theta, \theta'(0, \eta_C))$ is strictly increasing with respect to $\eta_C \in [0, 1/J(\theta))$ and it has a local maximum point at $\eta_C = 1/J(\theta)$, which completes the proof of the first half.

Next, we show the last half of the theorem. The derivative of $Q(\theta, \theta'(\eta_m, \eta_C))$ with respect to $\eta_m$ is
\begin{equation*}
\begin{split}
 \frac{\partial Q(\theta, \theta'(\eta_m, \eta_C))}{\partial \eta_m} &= \tilde \nabla_{\theta_m} J(\theta)^\mathrm{T} \mathbb{E}[f(\mathbf{x}) \nabla_{\theta_m} \ln \pi(\mathbf{x} \mid \theta'(\eta_m, \eta_C))] / J(\theta)\\
 &= \tilde \nabla_{\theta_m} J(\theta)^\mathrm{T} \mathbb{E}[f(\mathbf{x}) (\mathbf{C}_{\eta_C})^{-1} (\mathbf{x} - \mathbf{m}_{\eta_m})] / J(\theta)\\
 &= \tilde \nabla_{\theta_m} J(\theta)^\mathrm{T} (\mathbf{C}_{\eta_C})^{-1} \mathbb{E}[f(\mathbf{x}) (\mathbf{x} - \mathbf{m}) - \eta_m \tilde \nabla_{\theta_m} J(\theta)] / J(\theta)\\
 &= (1/J(\theta) - \eta_m) \tilde \nabla_{\theta_m} J(\theta)^\mathrm{T} (\mathbf{C}_{\eta_C})^{-1} \tilde \nabla_{\theta_m} J(\theta).
\end{split}
\end{equation*} 
Taking into account that $\mathbf{C}_{\eta_C}$ is positive definite for $\eta_C \in [0, 1/J(\theta)]$, it is easy to verify that $Q(\theta, \theta'(\eta_m, \eta_C))$ is strictly increasing with $\eta_m \in [0, 1/J(\theta)]$ and has the peak at $\eta_m = 1/J(\theta)$. This completes the proof.\qed
\end{proof}
% End Proof %%%%%%%%%%%%%%%%%%%%%%%%%%%%%%%%%%%%%%%%%%%%%%%%%%%%%%%%%%%%%%%%%%%

To provide an intuitive explanation of this theorem, we first show what happens at the maximum point. Let $\eta_m = \eta_C = 1/J(\theta^t)$. Then, according to Theorem~\ref{thm:1} we have
\begin{align}
 \mathbf{m}^{t+1} &= \int \frac{f(\mathbf{x}) \pi(\mathbf{x}; \theta^t) }{J(\theta^t)} \mathbf{x}\ d\mathbf{x},\label{eq:m-new}\\
 \mathbf{C}^{t+1} &= \int \frac{f(\mathbf{x}) \pi(\mathbf{x}; \theta^t) }{J(\theta^t)} (\mathbf{x}-\mathbf{m}^t)(\mathbf{x}-\mathbf{m}^t)^\mathrm{T}\ d\mathbf{x}.\label{eq:c-new}
\end{align}
That is, the past information is forgotten and the next estimates are only determined by the current information when the learning rates are taken so as to maximize the lower bound \eqref{eq:lower-bound}. 

\begin{figure}
\centering
\includegraphics[width=0.5\textwidth]{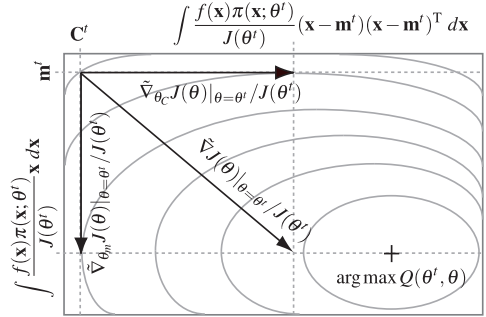}
\caption{The relation between the natural gradient of $J(\theta)$ at $\theta^{t}$, the target points, and the contour lines (solid gray curves) of $Q(\theta^{t}, \cdot)$.}
\label{fig:lower-bound}
\end{figure}

Now we restate Theorem~\ref{thm:2}. For large $\lambda$ such that the estimates \eqref{eq:ngl-cma-m} and \eqref{eq:ngl-cma-c} approximate the natural gradients sufficiently well, $\eta_m = \eta_C = 1/J(\theta^t)$ seems to be the best choice. Then, the next estimates become \eqref{eq:m-new} and \eqref{eq:c-new}. Therefore, the theorem says that moving the parameters toward \eqref{eq:m-new} and \eqref{eq:c-new} leads to increase of the expected fitness even when we assign different values to learning rates $\eta_m$ and $\eta_C$. Fig.~\ref{fig:lower-bound} illustrates the relation between the natural gradients, the target points, and $Q(\theta^t, \cdot)$.

\subsection{Justification of the Learning Rates in the CMA-ES}
\label{sec:improvement:cma}

Remembering that $\tilde\nabla J(\theta)/J(\theta) = \tilde\nabla \ln J(\theta)$ and that the update rules \eqref{eq:cma-m} and \eqref{eq:cma-c} in the CMA-ES are more similar to $\tilde\nabla \ln J(\theta)$ than $\tilde\nabla J(\theta)/J(\theta)$, which is mentioned in Section~\ref{sec:natural:cma}, Theorem~\ref{thm:2} justifies the constant and different learning rates in the CMA-ES: When $\lambda$ is large enough, it can be considered appropriate to set the learning rates to nearly one, because the lower bound \eqref{eq:lower-bound} of the increment in the log of expected fitness is maximized then. When $\lambda$ is not large enough, smaller learning rates seem to be appropriate to avert a fluctuation of parameters due to the large variance of natural gradient estimation. Since $\theta_m$ and $\theta_C$ have different sizes and the variances of the gradient estimates differ between $\mathbf{m}$-part and $\mathbf{C}$-part, it is natural to set the learning rates to different values. 

\subsection{Similarity to the EM-based Algorithm and Information Geometric Interpretation}
\label{sec:improvement:em}

From Theorem~\ref{thm:2}, we can view natural gradient learning for expected fitness maximization as an iterative method for finding the value of $\theta^{t+1}$ that improves $Q(\theta^{t}, \theta^{t+1})$ compared to $Q(\theta^{t}, \theta^{t})$. This is similar to the fitness expectation maximization \cite{WierstraPPSN2008}, whose framework is inspired by expectation maximization (EM) algorithms \cite{DempsterJRSS1977}. Here we discuss the relation to the EM-based algorithm to introduce an information geometric interpretation of the CMA-ES.

\paragraph{EM and EM-based Search Algorithms} In semi-supervised learning scenarios, EM algorithms seek to find a maximum-likelihood estimate of parameters of statistical models that depend on latent variables by alternating between an expectation (E) step and a maximization (M) step. The E-step calculates the expectation of the log-likelihood using the current estimate and the M-step finds the parameter that maximizes the expectation. In reinforcement learning \cite{DayanNC1997,KoberNIPS2009} and continuous optimization \cite{WierstraPPSN2008} scenario, EM based algorithms seek to find the optimal parameters that maximize expected reward or expected fitness by taking into account the inequality \eqref{eq:lower-bound}. The counterpart of E-step calculates the expectation $Q(\theta^t, \theta^{t+1})$ of the log-likelihood function $\ln\pi(\mathbf{x}; \theta^{t+1})$ under $q(\mathbf{x}; \theta^{t})$ defined previously. The counterpart of M-step finds the $\theta^{t+1}$ value that maximizes $Q(\theta^{t}, \theta^{t+1})$. The fitness expectation maximization algorithm constitutes an algorithm similar to the estimation of multivariate normal algorithm ($\text{EMNA}_\text{global}$; \cite{Larranaga2002book}), which is a variant of estimation of distribution algorithms (EDA).

\begin{figure}
\centering
\includegraphics[width=0.5\textwidth]{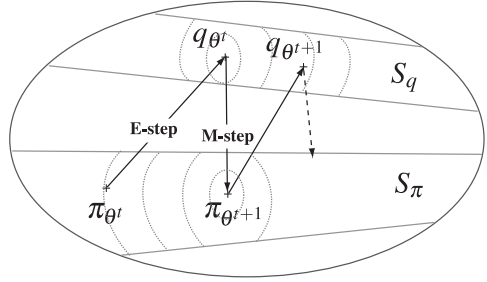}
\caption{Geometric Interpretation of the EM-based algorithm. Dotted gray curves represent the contour lines of KL divergence $D_\text{KL}(q_{\theta^{t}} \parallel \cdot)$ from $q_{\theta^{t}}$.}
\label{fig:em-geometric}
\end{figure}

\paragraph{Geometric View of the EM-based Algorithm} Let $S_{\pi} = \{ \pi_{\theta} = \pi(\mathbf{x}; \theta) \mid \theta \in \Theta \}$ and $S_{q} = \{ q_{\theta} = f(\mathbf{x})\pi(\mathbf{x}; \theta)/ J(\theta) \mid \theta \in \Theta\}$ be statistical manifolds. Considering the equality
\begin{align}
 Q(\theta^{t}, \theta^{t+1}) - Q(\theta^{t}, \theta^{t}) &= -H(q_{\theta^{t}}, \pi_{\theta^{t+1}}) + H(q_{\theta^{t}}, \pi_{\theta^{t}}) \notag\\
 &= -H(q_{\theta^{t}}, \pi_{\theta^{t+1}}) + H(q_{\theta^{t}}) + H(q_{\theta^{t}}, \pi_{\theta^{t}}) - H(q_{\theta^{t}}) \notag\\
 &= D_\mathrm{KL}(q_{\theta^{t}} \parallel \pi_{\theta^{t}}) - D_\mathrm{KL}(q_{\theta^{t}} \parallel \pi_{\theta^{t+1}}), \label{eq:kl-dif}
\end{align}
we find that choosing $\theta^{t+1}$ so that it maximizes $Q(\theta^{t}, \theta^{t+1})$ is equivalent to find $\pi_{\theta^{t+1}}$ on $S_{\pi}$ closest to current distribution $q_{\theta^{t}}$ on $S_{q}$ with respect to KL divergence. Based on the equality \eqref{eq:kl-dif} and the information geometry view of EM algorithms \cite{NealLGM1998,AmariNN1995}, we perceive the EM based algorithm as a repeated projection method between $S_{\pi}$ and $S_{q}$, where the projection corresponding to the E-step maps $\pi_{\theta^{t}}$ to $q_{\theta^{t}}$ and the projection corresponding to the M-step finds $\pi_{\theta^{*}} \in S_{\pi}$ that is the nearest from $q_{\theta^{t}}$ with respect to KL divergence (see Fig.~\ref{fig:em-geometric}). 

\paragraph{Information Geometry of the CMA-ES} The EM-based algorithm performs maximization of $D_\mathrm{KL}(q_{\theta^{t}} \parallel \pi_{\theta^{t}}) - D_\mathrm{KL}(q_{\theta^{t}} \parallel \pi_{\theta^{t+1}})$ in $\pi_{\theta^{t+1}}$, which is a lower bound of the expected fitness improvement, but the CMA-ES just moves the sampling distribution to a distribution on $S_\pi$ that is closer (not closest) to the target distribution $q_{\theta^{t}}$. This corresponds to generalized EM (GEM) algorithms \cite{DempsterJRSS1977} where the M-step is replaced with a step that finds the $\theta^{t+1}$ value that only improves the expected value. 

\begin{figure}
\centering
\includegraphics[width=0.5\textwidth]{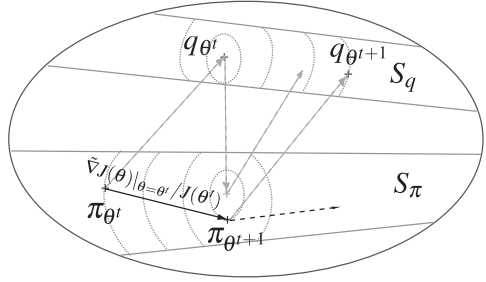}
\caption{Geometric Interpretation of the CMA-ES. Dotted gray curves represent the contour lines of KL divergence $D_\text{KL}(q_{\theta^{t}} \parallel \cdot)$ from $q_{\theta^{t}}$.}
\label{fig:cma-geometric}
\end{figure}

An important property and possibly an advantage of the CMA-ES over the EM-based algorithm is that the CMA-ES employs the natural gradient of the expected fitness $J(\cdot)$ itself. According to the equality
\begin{equation*}
\ln J(\theta^{t+1}) - \ln J(\theta^{t}) = D_\text{KL}(q_{\theta^{t}} \parallel q_{\theta^{t+1}}) + D_\mathrm{KL}(q_{\theta^{t}} \parallel \pi_{\theta^{t}}) - D_\mathrm{KL}(q_{\theta^{t}} \parallel \pi_{\theta^{t+1}}),
\end{equation*}
which is derived from equalities \eqref{eq:improvement} and \eqref{eq:kl-dif}, the improvement in the expected fitness is determined by both $D_\text{KL}(q_{\theta^{t}} \parallel q_{\theta^{t+1}})$ and $D_\mathrm{KL}(q_{\theta^{t}} \parallel \pi_{\theta^{t}}) - D_\mathrm{KL}(q_{\theta^{t}} \parallel \pi_{\theta^{t+1}})$. The CMA-ES moves the sampling distribution along the natural gradient of the expected fitness and turns out to make it closer to the target distribution. It does not perform maximization of the second amount but it also takes the first amount into account, whereas the EM-based algorithm maximizes the second amount but does not take the first amount into consideration (see Fig.~\ref{fig:cma-geometric}).

% Section 6 %%%%%%%%%%%%%%%%%%%%%%%%%%%%%%%%%%%%%%%%%%%%%%%%%%%%%%%%%%%%%%%%%%%
\section{Summary}
\label{sec:summary}

We described the analogy between the CMA-ES and natural gradient learning for (the log of) the expected fitness maximization in Section~\ref{sec:natural}. If one sets the weights in \eqref{eq:cma-m} and \eqref{eq:cma-c} to be $f(\mathbf{x}_{i})/\sum_{j=1}^{\lambda} f(\mathbf{x}_{j})$ at each iteration, adjustment of the parameters in the CMA-ES is equivalent to the estimate of the natural gradient of the log of expected fitness. In addition, the weights share some properties with practically used weights in the CMA-ES. Next, we investigated the properties of natural gradient learning in Section~\ref{sec:improvement}. We derived the range of learning rates that guarantee that the step along the exact natural gradient will increase the expected fitness and justified the use of different learning rates for each parameter. By considering the similarity to the EM-based algorithm, we showed that natural gradient learning with derived range of learning rates can be considered as a generalized EM-based algorithm. Natural gradient learning finds the parameters such that the sampling distribution $\pi(\mathbf{x}; \theta^{t+1})$ better matches the current target distribution $f(\mathbf{x})\pi(\mathbf{x}; \theta^{t}) / J(\theta^{t})$. However, in contrast to the EM-based algorithm, it does not minimize the divergence between the distributions but takes the other quantity contained in $J(\theta^t)$ into consideration. Finally, we provided an information geometry interpretation of the CMA-ES. 

Our results contribute to the theoretical aspect of the CMA-ES and to the improvement of the CMA-ES. The natural gradient view together with the EM like view will help to construct the convergence (stability) theory of the CMA-ES. Information geometry view might give some insight into more efficient and effective parameter updates. %These would be future work.

In this paper, we did not treat the evolution paths. As we mentioned in Section~\ref{sec:cma}, they have a great impact on the performance when $\lambda$ is small. A theoretical foundation for the evolution paths is desired. In addition, we did not consider the inaccuracy of the natural gradient estimation. We analyze the stability of the CMA-ES in the future work. Furthermore, as mentioned in Section~\ref{sec:natural:cma}, further investigation about fitness shaping, i.e. the coefficients in the natural gradient estimation, is also an important future work. 

%\bibliographystyle{amsalpha}
%\bibliography{PublishedReferences,library,../../../paperslib}

\end{document}